\documentclass[lettersize, journal]{IEEEtran}
\usepackage{amsmath,amsfonts}
\usepackage[noend]{algorithmic}
\usepackage{algorithm}
\usepackage{array}
\usepackage[caption=false,font=normalsize,labelfont=sf,textfont=sf]{subfig}
\usepackage{textcomp}
\usepackage{stfloats}
\usepackage{url}
\usepackage{verbatim}
\usepackage{graphicx}
\usepackage{cite}

\usepackage{xcolor}
\def\BibTeX{{\rm B\kern-.05em{\sc i\kern-.025em b}\kern-.08em
    T\kern-.1667em\lower.7ex\hbox{E}\kern-.125emX}}

\usepackage{booktabs}
\usepackage{multirow}
\usepackage[super]{nth}
\usepackage{tikz}
\newcommand*\circled[1]{\tikz[baseline=(char.base)]{
		\node[shape=circle,draw,inner sep=0.2pt] (char) {#1};}}
\newcommand*\circledB[1]{\tikz[baseline=(char.base)]{
            \node[shape=circle,fill,inner sep=0.2pt] (char) {\textcolor{white}{#1}};}}
\usepackage{soul}
\usepackage{enumitem}
\usepackage{setspace}
\newcolumntype{?}{!{\vrule width 1.5pt}}

\usepackage{hyphenat}
\usepackage{fancyhdr}
\fancypagestyle{firstpage}{
  \fancyhf{}
  \chead{Accepted for publication at the IEEE Embedded Systems Letters}
  \cfoot{\thepage}
}

\begin{document}

\title{SpiKernel: A Kernel Size Exploration Methodology for Improving Accuracy of the Embedded \\ Spiking Neural Network Systems}

\author{Rachmad Vidya Wicaksana Putra,~\IEEEmembership{Member,~IEEE,} and Muhammad Shafique,~\IEEEmembership{Senior Member,~IEEE} 
\thanks{Rachmad Vidya Wicaksana Putra is with eBrain Lab, Division of Engineering, New York University (NYU) Abu Dhabi, United Arab Emirates;
{e-mail: rachmad.putra@nyu.edu}}
\thanks{Muhammad Shafique is the Director of eBrain Lab, Division of Engineering, New York University (NYU) Abu Dhabi, United Arab Emirates;
{e-mail: muhammad.shafique@nyu.edu}}
\thanks{Manuscript received Month DD, 2024; revised Month DD, 2024.}
\vspace{-0.9cm}
}

\maketitle
\pagestyle{plain}
\thispagestyle{firstpage}

\begin{abstract}
Spiking Neural Networks (SNNs) can offer ultra-low power/energy consumption for machine learning-based application tasks due to their sparse spike-based operations. 
Currently, most of the SNN architectures need a significantly larger model size to achieve higher accuracy, which is not suitable for resource-constrained embedded applications. 
Therefore, developing SNNs that can achieve high accuracy with acceptable memory footprint is highly needed. 
Toward this, we propose \textit{SpiKernel}, a novel methodology that improves the accuracy of SNNs through kernel size exploration.   
Its key steps include (1) investigating the impact of different kernel sizes on the accuracy, (2) devising new sets of kernel sizes, (3) generating SNN architectures using neural architecture search based on the selected kernel sizes, and (4) analyzing the accuracy-memory trade-offs for SNN model selection.
The experimental results show that our SpiKernel achieves higher accuracy than state-of-the-art works (i.e., 93.24\% for CIFAR10, 70.84\% for CIFAR100, and 62\% for TinyImageNet) with less than 10M parameters and up to 4.8x speed-up of searching time, thereby making it suitable for embedded applications.
\end{abstract}

\begin{IEEEkeywords}
Spiking neural networks, kernel size exploration, neural architecture search, embedded applications.
\end{IEEEkeywords}

\vspace{-0.3cm}
\section{Introduction}
\label{Sec_Intro}

\begin{figure}[t]
\centering
\includegraphics[width=\linewidth]{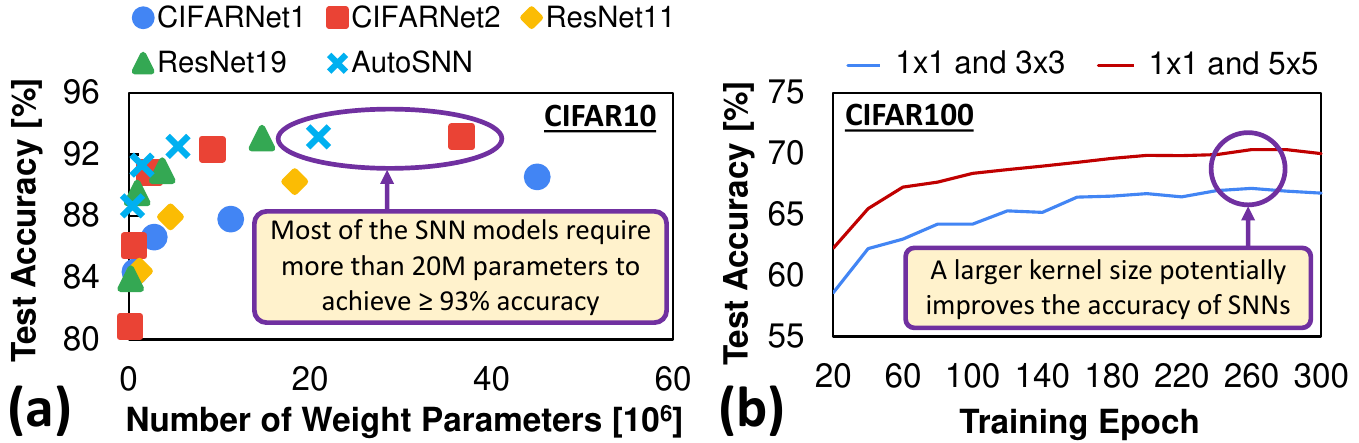}
\vspace{-0.7cm}
\caption{(a) Accuracy and memory footprints of different SNNs for the CIFAR10: AutoSNN~\cite{Ref_Na_AutoSNN_ICML22},
ResNet11~\cite{Ref_Lee_SpikeBackprop_FNINS20},
ResNet19~\cite{Ref_Zheng_LargerSNNs_AAAI21},  CIFARNet1~\cite{Ref_Wu_DirectTrainSNNs_AAAI19}, CIFARNet2~\cite{Ref_Fang_MemTConstantSNNs_ICCV21}; adapted from the studies in~\cite{Ref_Na_AutoSNN_ICML22}. 
(b) Experimental results considering the CIFAR100 dataset and two SNN architectures with different sets of kernel sizes: one architecture with 1x1 and 3x3, and another one with 1x1 and 5x5. This shows that a larger kernel size may improve SNN accuracy.} 
\label{SNN_AccMem_CaseStudy}
\vspace{-0.6cm}
\end{figure}

Recent studies have shown that SNN algorithms can solve many machine learning (ML) applications, such as computer vision~\cite{Ref_Putra_FSpiNN_TCAD20}\cite{Ref_Putra_QSpiNN_IJCNN21},  robotics~\cite{Ref_Bing_RoboticsSNNs_FNBOT18}\cite{Ref_Putra_NeuromorphicAI4Robotics_arXiv24}, automotive~\cite{Ref_Cordone_ObjDetSNN_IJCNN22}\cite{Ref_Putra_SNN4Agents_FROBT24}, 
and healthcare~\cite{Ref_Luo_EEGSNN_Access20}, while incurring ultra-low power/energy consumption.  
Currently, the existing SNN models still consume huge memory footprints to achieve high accuracy. 
For instance, to achieve at least 93\% accuracy for the CIFAR10 dataset, most of SNN models require more than 20x10$^6$ (20M) parameters as shown in Fig.~\ref{SNN_AccMem_CaseStudy}(a), which is not suitable for resource-constrained embedded applications.
Therefore, \textit{\textbf{the targeted research problem} is how can we develop SNN architectures that achieve high accuracy with acceptable memory footprints for resource-constrained embedded applications}. 

Developing an SNN architecture using a widely-used conversion method from a non-spiking Artificial Neural Network (ANN) to an SNN is not suitable for the targeted problem, since the SNN model size will depend on the original ANN model size, which may not satisfy the memory constraints.
Moreover, a recent work~\cite{Ref_Kim_SNASNet_ECCV22} demonstrated that developing SNN architectures directly in spiking domain can achieve higher accuracy than the conversion method. 
However, the state-of-the-art typically still employ a set of small convolution kernel sizes (i.e., 1x1 and 3x3)~\cite{Ref_Kim_SNASNet_ECCV22}\cite{Ref_Na_AutoSNN_ICML22}, which limit their feature extraction capabilities and lead to sub-optimal accuracy.
To show these limitations, we perform an experimental case study to evaluate the accuracy of SNNs with different sets of kernel sizes. 
Here, we generate two SNN architectures using neural architecture search (NAS) concept from~\cite{Ref_Kim_SNASNet_ECCV22}: (1) an SNN that is generated using the 1x1 and 3x3 kernels to represent small kernel sizes, and (2) an SNN that is generated using the 1x1 and 5x5 kernels to represent large kernel sizes.
Note, the NAS concept is explained in Section~\ref{Sec_Back_NAS}.
The experimental results are presented in Fig.~\ref{SNN_AccMem_CaseStudy}(b), and these results indicate that employing a larger kernel size (i.e., 5x5) can improve the accuracy due to its better spatial feature extraction than the smaller kernel size (i.e., 3x3).   
\textit{This observation exposes the opportunity to improve SNN accuracy by increasing the kernel size}. 
However, simply increasing the kernel size without a systematic approach may significantly increase the memory footprint without notable accuracy improvement.  
Therefore, \textit{\textbf{the research challenge} is how to decide the appropriate kernel sizes for achieving high accuracy while incurring acceptable memory footprint for resource-constrained embedded systems}. 

\textbf{Our Novel Contributions:}
To address the research problem and related challenge, \textit{we propose \textbf{SpiKernel}, a novel methodology to explore kernel sizes for improving the accuracy of SNNs, while considering the memory footprint}. 
It is achieved by investigating the impact of different kernel sizes, devising new sets of kernel sizes, generating SNN architectures based on the new sets of kernel sizes, and analyzing the accuracy-memory trade-offs for network model selection.

\vspace{-0.3cm}
\section{Background of NAS for SNNs}
\label{Sec_Back_NAS}

\begin{figure}[hbtp]  
\vspace{-0.3cm}
\centering
\includegraphics[width=\linewidth]{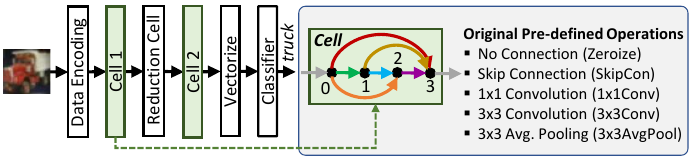}
\vspace{-0.7cm}
\caption{The SNN macro-architecture, employing 2 neural cells; where a neural cell is defined as a directed acyclic graph with each connection edge represents a specific pre-defined operation; adapted from~\cite{Ref_Kim_SNASNet_ECCV22}.} 
\label{Fig_MacroArch}
\vspace{-0.2cm}
\end{figure}

Manually developing an SNN architecture is time consuming and laborious~\cite{Ref_Putra_SpikeNAS_arXiv24}, hence the employment of NAS concept for SNN generation is necessary. 
Toward this, we leverage the cell-based NAS concept to provide a unified benchmark and ensure reproducibility for NAS algorithms~\cite{Ref_Dong_NASbench201_ICLR20}.
The state-of-the-art cell-based NAS for SNNs is SNASNet~\cite{Ref_Kim_SNASNet_ECCV22}, whose concept is employing a macro-architecture that has 2 neural cells; see Fig.~\ref{Fig_MacroArch}. 
Each cell consists of 4 nodes and edges. 
Each edge refers to a specific operation from the pre-defined operations: \textit{no connection} (Zeroize), \textit{skip connection} (SkipCon), \textit{1x1 convolution} (1x1Conv), \textit{3x3 convolution} (3x3Conv), and \textit{3x3 average pooling} (3x3AvgPool).
In the searching process, different combinations of operations in the cell are explored to find the architecture that can achieve high accuracy. 

\textit{In this work, we follow the NAS technique from SNASNet~\cite{Ref_Kim_SNASNet_ECCV22} with feed-forward topology to evaluate the impact of different kernel sizes and generate the respective SNN architectures}.
In this manner, we combine the strengths of the kernel size design and the cell-based NAS concept to ensure the generality of our proposed methodology.
This NAS employs a random search mechanism based on the defined number of search iteration (i.e., 5000x iteration) to determine the cell architecture in the network.    
Furthermore, this NAS also avoids the training process for evaluating the network, rather it quickly measures the representation capability score ($R$) of an investigated architecture for achieving high accuracy at the early stage through the following key steps.
\begin{enumerate}
    \item Input samples in a mini-batch are fed to the network, and the neuron activities in each layer are recorded as binary vector $f$. 
    \item A matrix $\textbf{K}$ for each layer is formed by calculating the Hamming distance $H(f_i, f_j)$ between samples $i$ and $j$. 
    \item The $R$ score of the network is calculated using Eq.~\ref{Eq_Kscore}. Afterward, the network with the highest $R$ is selected as a network candidate and ready to train.
\end{enumerate}
\vspace{0.3cm}
\begin{equation}
  \centering
  \begin{split}
    R = \log (\det \; \lvert \sum_l \textbf{K}^l \rvert)
    \;\;\; \text{with} \; l \; \text{is the network layer-}l
    \end{split}
    \label{Eq_Kscore}
\end{equation}

\section{Our SpiKernel Methodology}
\label{Sec_Method}
\vspace{-0.3cm}

\textbf{\begin{figure}[t]
\centering
\includegraphics[width=0.9\linewidth]{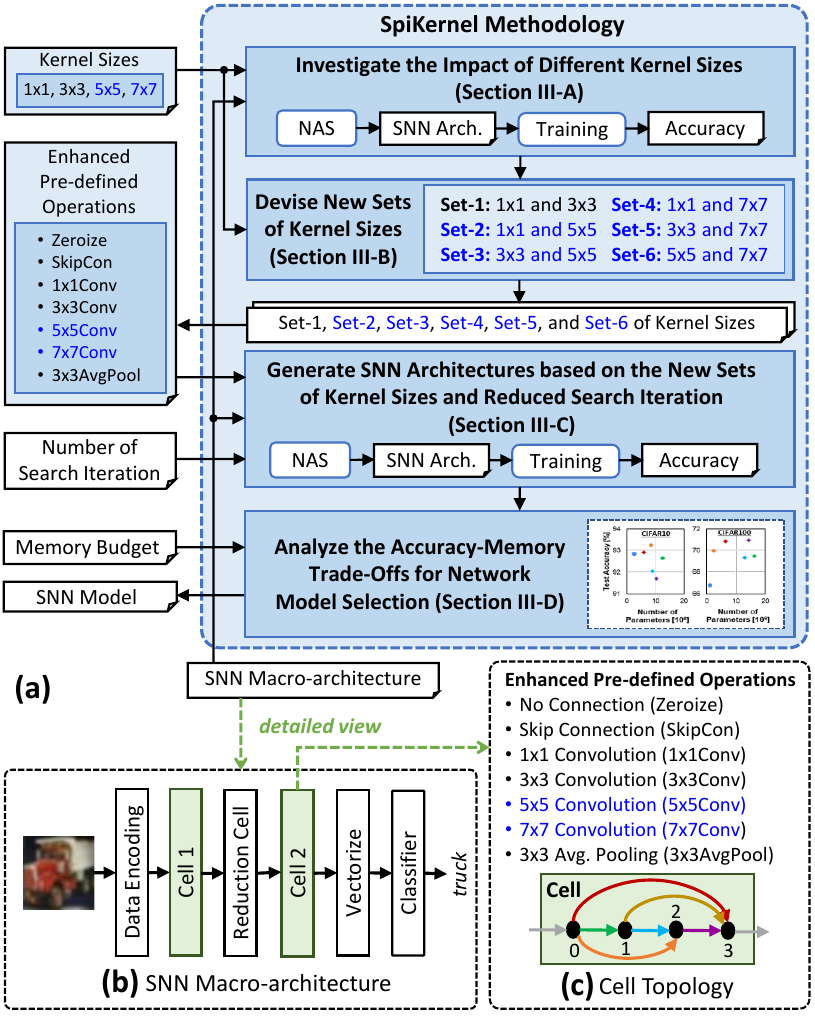}
\vspace{-0.3cm}
\caption{(a) Our SpiKernel methodology with novel contributions highlighted in blue boxes. 
(b) The SNN macro-architecture considered in the SpiKernel. 
(c) The neural cell in the SpiKernel, showing the enhanced pre-defined operations with additional kernel sizes and operations in \textit{`blue text'} as compared to the original ones shown in Fig.~\ref{Fig_MacroArch}.} 
\label{Fig_SpiKernel_Method}
\vspace{-0.3cm}
\end{figure}}

To address the targeted problem and related challenge, our SpiKernel methodology employs several key steps; see Fig.~\ref{Fig_SpiKernel_Method}. 
Their details are discussed in Section~\ref{Sec_Method_Investigate} to Section~\ref{Sec_Method_TradeOffs}.

\vspace{-0.3cm}
\subsection{Investigating the Impact of Different Kernel Sizes}
\label{Sec_Method_Investigate}

\begin{figure}[t]
\centering
\includegraphics[width=\linewidth]{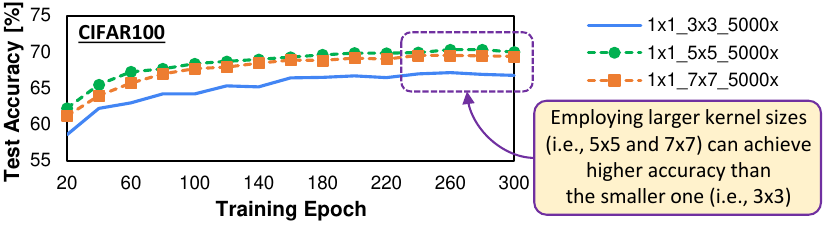}
\vspace{-0.7cm}
\caption{Results of experimental case studies using NAS for the CIFAR100 dataset, while considering different sets of kernel sizes: 1x1Conv and 3x3Conv (i.e., 1x1\_3x3\_5000x), 1x1Conv and 5x5Conv (i.e., 1x1\_5x5\_5000x), as well as 1x1Conv and 7x7Conv (i.e., 1x1\_7x7\_5000x).} 
\label{Fig_CaseStudy}
\vspace{-0.5cm}
\end{figure}

This step aims at understanding if employing a larger kernel size can improve the accuracy from the smaller ones.
For this, we employ NAS technique from Section~\ref{Sec_Back_NAS} to generate an SNN architecture based on the given set of kernel sizes. 
In this work, we consider the 1x1, 3x3, 5x5, and 7x7 kernel sizes. 
The 1x1 kernel aims at minimizing the computational requirements via point-wise convolution (1x1Conv)~\cite{Ref_Howard_MobileNet_arXiv17}. 
Meanwhile, the 3x3, 5x5, and 7x7 kernels aim at extracting input features through convolution operations (i.e., 3x3Conv, 5x5Conv, and 7x7Conv, respectively). 
These kernel sizes are combined into several sets: (1x1Conv and 3x3Conv), (1x1Conv and 5x5Conv), and (1x1Conv and 7x7Conv), which are then incorporated into the NAS process.
Afterward, the generated SNN architecture is trained and tested subsequently to evaluate the accuracy, and the experimental results are presented in Fig.~\ref{Fig_CaseStudy}. 
These results indicate that \textit{in general, a larger kernel size can improve the accuracy from the smaller ones}, because it can capture more information for unique feature extraction. 

\vspace{-0.3cm}
\subsection{Devising New Sets of Kernel Sizes for NAS}
\label{Sec_Method_KernelSizes}

This step aims to define the kernel sizes (and the respective operations) for developing a network architecture. 
Therefore, we develop several new kernel sets based on the kernel sizes explored in Section~\ref{Sec_Method_Investigate}, in addition to the typical kernel set employed in the state-of-the-art (i.e., 1x1 and 3x3)~\cite{Ref_Kim_SNASNet_ECCV22}.
Specifically, we combine 2 different kernels from the user-defined kernel sizes (i.e., 1x1, 3x3, 5x5, and 7x7) to introduce variability of feature extraction capabilities and accuracy.
As a result, we obtain 6 kernel sets, namely Set-1 (1x1 and 3x3), Set-2 (1x1 and 5x5), Set-3 (3x3 and 5x5), Set-4 (1x1 and 7x7), Set-5 (3x3 and 7x7), and Set-6 (5x5 and 7x7), as shown in Table~\ref{Tab_KernelSets}.
\textit{These sets of kernel sizes will be employed by NAS process for exploring SNN architecture candidates that can offer high accuracy with acceptable memory footprints}.

\begin{table}[hbtp]
\vspace{-0.3cm}
\caption{New Sets of Kernel Sizes for NAS in SpiKernel.}
\vspace{-0.2cm}
\centering
\footnotesize
\begin{tabular}{|c|c?c|c|c|} 
\hline
\textbf{Set} & \textbf{Kernel Sizes} & \textbf{Set} & \textbf{Kernel Sizes} \\
\hline 
\hline
Set-1 & (1x1Conv, 3x3Conv) & Set-2 & (1x1Conv, 5x5Conv) \\
\hline
Set-3 & (3x3Conv, 5x5Conv) & Set-4 & (1x1Conv, 7x7Conv) \\
\hline
Set-5 & (3x3Conv, 7x7Conv) & Set-6 & (5x5Conv, 7x7Conv) \\
\hline
\end{tabular}
\label{Tab_KernelSets}
\vspace{-0.2cm}
\end{table}

Note, these new kernel sets are manually crafted. 
Therefore, users can define their own kernel sets based on their design specifications.
For instance, users may consider a fewer kernel sets (less than 6 kernel sets) to further expedite the exploration process.
In this manner, our SpiKernel methodology is highly flexible for facilitating any given kernel sets, datasets, and memory constraints to generate appropriate SNN architectures.

\vspace{-0.3cm}
\subsection{Generating SNN Architectures based on the New Sets of Kernel Sizes and Reduced Search Iteration}
\label{Sec_Method_GenerateSNN}
\vspace{-0.1cm}

This step aims at developing SNN architectures based on the new sets of kernel sizes, while optimizing the searching time. 
To do this, we leverage NAS technique from Section~\ref{Sec_Back_NAS} and kernel sets from Table~\ref{Tab_KernelSets} to generate SNN architectures. 
To optimize the searching time, we consider reducing the number of random search iteration from the original setting since previous work~\cite{Ref_Putra_SpikeNAS_arXiv24} has observed that it is possible to maintain accuracy with a smaller number of random search iteration.   
Specifically, we employ 1000x search iteration, which is smaller than the state-of-the-art work (i.e., 5000x search iteration). 
For each given set of kernel sizes, NAS process generates a network candidate.
Then, each network candidate is evaluated for its accuracy and memory size. 
 
\vspace{-0.3cm}
\subsection{Accuray-Memory Trade-Off Analyses for Model Selection}
\label{Sec_Method_TradeOffs}
\vspace{-0.1cm}

Different SNN architectures that are obtained from NAS in Section~\ref{Sec_Method_GenerateSNN} have different memory requirements, hence they might not satisfy the memory constraints from the target applications.
Toward this, we analyze the accuracy-memory trade-offs to select the appropriate SNN models. 
To do this, we employ the Alg.~\ref{Alg_ModelSelection} to select the most suitable SNN model from the existing candidates.
Its key idea is to save the SNN model that achieves higher accuracy, while meeting the memory constraint. 
If multiple candidates are applicable, then the model that achieves the highest accuracy is selected.

To evaluate the memory footprint of a network, we leverage the number of network parameters ($N$), which can be calculated using Eq.~\ref{Eq_Param1}-\ref{Eq_Param2}, following the studies in \cite{Ref_Putra_SpikeNAS_arXiv24}. 
$N_{w}^l$ and $N_b^l$ represent the number of weights and bias in layer-$l$, respectively. 
The dimension of weight filters in the given layer is represented with $h_1$ for height, $h_2$ for width, $c$ for the number of channel, and $r$ for the number of filters. 
\begin{equation}
  \begin{split}
    N = \sum_l (N_{w}^l+N_b^l) 
    \end{split}
    \label{Eq_Param1}
  \vspace{-0.5cm}
\end{equation}
\begin{equation}
  \begin{split}
    N_{w} = h_1 \cdot h_2 \cdot c \cdot r
    \end{split}
    \label{Eq_Param2}
  \vspace{-0.5cm}
\end{equation}

\vspace{-0.4cm}
\section{Evaluation Methodology}
\label{Sec_EvalMethod}

We evaluate our SpiKernel methodology through a Python-based implementation that runs on the Nvidia RTX 4090 GPU machines. 
The network setting is shown as $A$x$A$\_$B$x$B$\_$I$x, which represents a network that is obtained through NAS process considering kernel sizes of $A$x$A$ and $B$x$B$, as well as $I$x search iteration. 
The state-of-the-art work employs the setting of 1x1\_3x3\_5000x, while our SpiKernel considers several settings: 1x1\_3x3\_1000x, 1x1\_5x5\_1000x, 3x3\_5x5\_1000x, 1x1\_7x7\_1000x, 3x3\_7x7\_1000x, and 5x5\_7x7\_1000x. 
Here, the inference configuration includes the employments of \textit{leaky integrate-and-fire (LIF) neuron model} to realize firing strategy and neuronal dynamics, \textit{rate coding} to perform the data/input encoding, \textit{timestep}=5 to define the operational time for a neuron to process the incoming spike train, and \textit{single-precision 32-bit floating point} to represent each weight value.

\begin{algorithm}[t]
\caption{Pseudo-code of the SNN model selection}
\label{Alg_ModelSelection}
\footnotesize
\begin{algorithmic}[1]
\renewcommand{\algorithmicrequire}{\textbf{INPUT:}}
\renewcommand{\algorithmicensure}{\textbf{OUTPUT:}}
\REQUIRE \textbf{(1)} Trained model candidates ($candidates$); \\
         \textbf{(2)} Memory constraint ($mem_{const}$); \\
         \textbf{(3)} Minimum acceptable accuracy ($acc_{th}$); \\
\ENSURE Selected SNN model ($model_{sel}$); \\
\smallskip
\textbf{BEGIN} \\
  \textbf{Initialization}: \\
  \STATE $model_{saved} = []$; \\
  \STATE $acc_{saved} = []$; \\
  \textbf{Process}: \\
    \FOR{($i=0$; $i<$ num($candidates$); $i$++)}
      \STATE $mem_{temp} =$ eval\_mem($candidates[i]$); // evaluate the memory\\
      \STATE $acc_{temp} =$ eval\_acc($candidates[i]$); // evaluate the accuracy\\
      \IF{($mem_{temp}$ $\leq$ $mem_{const}$) \AND ($acc_{temp}$ $\geq$ $acc_{th}$)}
          \STATE $model_{saved}[i] = candidates[i]$;\\
          \STATE $acc_{saved}[i] = acc_{temp}$;\\
      \ENDIF
    \ENDFOR
    \STATE $x =$ idx\_acc($acc_{saved}[i]$); // array index for the highest accuracy \\
    \STATE $model_{sel} = model_{saved}[x]$; // model selection
\RETURN $model_{sel}$; \\
\textbf{END}
\end{algorithmic} 
\end{algorithm}
\setlength{\textfloatsep}{4pt}

\vspace{-0.5cm}
\section{Results and Discussion}
\label{Sec_Results}
\vspace{-0.1cm}

\vspace{-0.2cm}
\subsection{Accuracy Improvements}
\label{Sec_Results_Accuracy}
\vspace{-0.2cm}

\begin{figure*}[t]
\centering
\includegraphics[width=\linewidth]{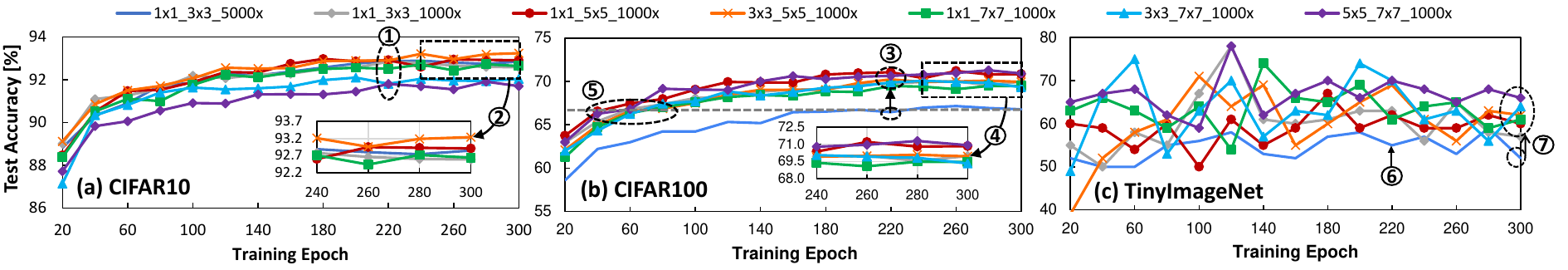}
\vspace{-0.8cm}
\caption{Experimental results for the test accuracy of different SNNs on (a) CIFAR10, (b) CIFAR100, and (c) TinyImageNet datasets.} 
\label{Fig_Results_Accuracy}
\vspace{-0.5cm}
\end{figure*}

Experimental results of the accuracy are shown in Fig.~\ref{Fig_Results_Accuracy}. 
For the CIFAR10, accuracy scores of the networks with larger kernel sizes are comparable to the state-of-the-art, as they are within 1\% accuracy range from the state-of-the-art; see \circled{1}. 
For instance, the network with 3x3\_5x5\_1000x achieves 93.24\% accuracy after 300 training epoch while the state-of-the-art achieves 92.85\% accuracy; see \circled{2}. 
For the CIFAR100, accuracy scores of the networks with larger kernel sizes are significantly higher than the state-of-the-art across all training epochs; see \circled{3}. 
For instance, the network with 1x1\_5x5\_1000x achieves 70.84\% accuracy after 300 training epoch while the state-of-the-art achieves 66.78\% accuracy; see \circled{4}. 
Furthermore, we can also achieve iso-accuracy like the state-of-the-art using a fewer training epochs (e.g., 60-100 epochs); see \circled{5}.
For the TinyImageNet (which has 200 classes), accuracy scores of the networks with larger kernel sizes are generally higher than the state-of-the-art; see \circled{6}-\circled{7}. 
The reason for all of these accuracy improvements is that, the larger kernel sizes can extract more unique input features (i.e., information) for generating a unique output feature. 
It is especially beneficial for tasks with large number of classes, as they have large number of unique features to distinguish. 
Therefore, the CIFAR100 and TinyImageNet tasks can benefit more from larger kernel sizes than the CIFAR10, as indicated by its notable accuracy improvements from the-state-of-the-art across all training epochs.

\subsection{Reduced Searching Time}
\label{Sec_Results_SearchTime}
\vspace{-0.1cm}

\begin{figure}[h]
\vspace{-0.3cm}
\centering
\includegraphics[width=\linewidth]{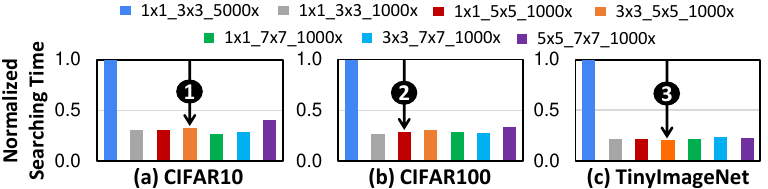}
\vspace{-0.7cm}
\caption{Experimental results for the searching time of different SNNs on (a) CIFAR10, (b) CIFAR100, and (c) TinyImageNet datasets.} 
\label{Fig_Results_SearchTime}
\vspace{-0.4cm}
\end{figure}

Experimental results of the searching time are shown in Fig.~\ref{Fig_Results_SearchTime}. 
These results show that our methodology leads to a faster searching time than the state-of-the-art, since it employs a smaller number of search iteration (1000x) than the state-of-the-art (5000x), while preserving the high accuracy. 
Compared to the state-of-the-art, the generated networks achieve about 3x searching time speed-up for CIFAR10 (with 3x3\_5x5\_1000x setting; see \circledB{1}), 3.45x speed-up for CIFAR100 (with 1x1\_5x5\_1000x setting; see \circledB{2}), and 4.8x speed-up for TinyImageNet (with 3x3\_5x5\_1000x setting; see \circledB{3}). 

\vspace{-0.3cm}
\subsection{Accuracy-Memory Trade-Offs}
\label{Sec_Results_TradeOffs}
\vspace{-0.1cm}

\begin{figure}[hbtp]
\vspace{-0.3cm}
\centering
\includegraphics[width=0.95\linewidth]{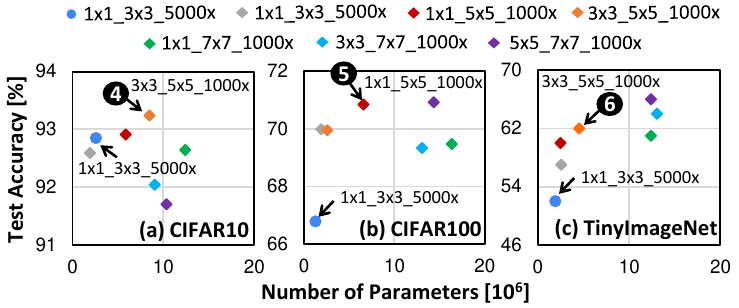}
\vspace{-0.4cm}
\caption{Experimental results for the accuracy-memory trade-offs of different SNNs on (a) CIFAR10, (b) CIFAR100, and (c) TinyImageNet datasets.} 
\label{Fig_Results_TradeOffs}
\vspace{-0.4cm}
\end{figure}

Correlation graphs between accuracy and memory footprint are shown in Fig.~\ref{Fig_Results_TradeOffs}.
These graphs show that networks with larger kernel sizes usually offer higher accuracy than smaller ones, while incurring larger memory footprints. 
To find the appropriate network for the given task, we apply the memory constraint on the correlation graph, and then select the one that meets this constraint, while offering high accuracy.  
For instance, if the application imposes 10$^6$ (or 10M) parameters as the memory budget, then we can select the network with 3x3\_5x5\_1000x for the CIFAR10 (see \circledB{4}), the one with 1x1\_5x5\_1000x for CIFAR100 (see \circledB{5}), and the one with 3x3\_5x5\_1000x for TinyImageNet (see \circledB{6}).  

\vspace{-0.3cm}
\subsection{Further Discussion}
\label{Sec_Results_Discuss}
\vspace{-0.1cm}

The state-of-the-art, that studied the impact of kernel sizes in SNNs~\cite{Ref_Lee_DSCNN_TCDS19}, only considered small datasets (i.e., MNIST with 10 classes and 28x28 input dimension), while employing user-crafted network architectures. 
In contrast, our SpiKernel methodology scales well across different complexity levels of datasets: CIFAR10 (10 classes and 32x32 input), CIFAR100 (100 classes and 32x32 input), and TinyImageNet (200 classes and 64x64 input).
Therefore, its computational efficiency and power consumption are not comparable to the state-of-the-art.
In summary, all these results demonstrated that our SpiKernel methodology effectively leverages any given kernel sizes to generate an SNN architecture that maximizes accuracy for any given dataset, while meeting the memory constraint.

\vspace{-0.2cm}
\section{Conclusion}
\label{Sec_Conclusion}
\vspace{-0.1cm}

We propose a novel SpiKernel methodology that improves SNN accuracy through kernel size exploration, i.e., by investigating the impact of kernel sizes, devising the new kernel sets, generating SNN architectures accordingly, and analyzing accuracy-memory trade-offs. 
The experimental results show that our SpiKernel offers higher accuracy than state-of-the-art with acceptable memory and reduced searching time, hence enabling efficient SNN deployment for embedded applications. 

\vspace{-0.2cm}
\begin{spacing}{1}
\bibliographystyle{IEEEtran}
\bibliography{bibliography}

\begin{thebibliography}{10}
\providecommand{\url}[1]{#1}
\csname url@samestyle\endcsname
\providecommand{\newblock}{\relax}
\providecommand{\bibinfo}[2]{#2}
\providecommand{\BIBentrySTDinterwordspacing}{\spaceskip=0pt\relax}
\providecommand{\BIBentryALTinterwordstretchfactor}{4}
\providecommand{\BIBentryALTinterwordspacing}{\spaceskip=\fontdimen2\font plus
\BIBentryALTinterwordstretchfactor\fontdimen3\font minus \fontdimen4\font\relax}
\providecommand{\BIBforeignlanguage}[2]{{%
\expandafter\ifx\csname l@#1\endcsname\relax
\typeout{** WARNING: IEEEtran.bst: No hyphenation pattern has been}%
\typeout{** loaded for the language `#1'. Using the pattern for}%
\typeout{** the default language instead.}%
\else
\language=\csname l@#1\endcsname
\fi
#2}}
\providecommand{\BIBdecl}{\relax}
\BIBdecl

\bibitem{Ref_Putra_FSpiNN_TCAD20}
R.~V.~W. {Putra} and M.~{Shafique}, ``Fspinn: An optimization framework for memory-efficient and energy-efficient spiking neural networks,'' \emph{IEEE Trans. on Computer-Aided Design of Integrated Circuits and Systems (TCAD)}, vol.~39, no.~11, pp. 3601--3613, 2020.

\bibitem{Ref_Putra_QSpiNN_IJCNN21}
R.~V.~W. Putra and M.~Shafique, ``Q-spinn: A framework for quantizing spiking neural networks,'' in \emph{IJCNN}, 2021, pp. 1--8.

\bibitem{Ref_Bing_RoboticsSNNs_FNBOT18}
Z.~Bing \emph{et~al.}, ``A survey of robotics control based on learning-inspired spiking neural networks,'' \emph{Frontiers in Neurorobotics}, vol.~12, 2018.

\bibitem{Ref_Putra_NeuromorphicAI4Robotics_arXiv24}
R.~V.~W. Putra \emph{et~al.}, ``Embodied neuromorphic artificial intelligence for robotics: Perspectives, challenges, and research development stack,'' \emph{arXiv preprint arXiv:2404.03325}, 2024.

\bibitem{Ref_Cordone_ObjDetSNN_IJCNN22}
L.~Cordone \emph{et~al.}, ``Object detection with spiking neural networks on automotive event data,'' in \emph{IJCNN}, 2022, pp. 1--8.

\bibitem{Ref_Putra_SNN4Agents_FROBT24}
R.~V.~W. Putra, A.~Marchisio, and M.~Shafique, ``Snn4agents: A framework for developing energy-efficient embodied spiking neural networks for autonomous agents,'' \emph{Frontiers in Robotics and AI}, vol.~11, 2024.

\bibitem{Ref_Luo_EEGSNN_Access20}
Y.~Luo \emph{et~al.}, ``Eeg-based emotion classification using spiking neural networks,'' \emph{IEEE Access}, vol.~8, pp. 46\,007--46\,016, 2020.

\bibitem{Ref_Kim_SNASNet_ECCV22}
Y.~Kim \emph{et~al.}, ``Neural architecture search for spiking neural networks,'' in \emph{ECCV}, 2022, pp. 36--56.

\bibitem{Ref_Na_AutoSNN_ICML22}
B.~Na \emph{et~al.}, ``Autosnn: Towards energy-efficient spiking neural networks,'' in \emph{ICML}, 2022, pp. 16\,253--16\,269.

\bibitem{Ref_Lee_SpikeBackprop_FNINS20}
C.~Lee \emph{et~al.}, ``Enabling spike-based backpropagation for training deep neural network architectures,'' \emph{Frontiers in Neuroscience}, p. 119, 2020.

\bibitem{Ref_Zheng_LargerSNNs_AAAI21}
H.~Zheng \emph{et~al.}, ``Going deeper with directly-trained larger spiking neural networks,'' in \emph{AAAI}, vol.~35, no.~12, 2021, pp. 11\,062--11\,070.

\bibitem{Ref_Wu_DirectTrainSNNs_AAAI19}
Y.~Wu \emph{et~al.}, ``Direct training for spiking neural networks: Faster, larger, better,'' in \emph{AAAI}, vol.~33, no.~01, 2019, pp. 1311--1318.

\bibitem{Ref_Fang_MemTConstantSNNs_ICCV21}
W.~Fang \emph{et~al.}, ``Incorporating learnable membrane time constant to enhance learning of spiking neural networks,'' in \emph{ICCV}, 2021, p. 2661.

\bibitem{Ref_Putra_SpikeNAS_arXiv24}
R.~V.~W. Putra and M.~Shafique, ``Spikenas: A fast memory-aware neural architecture search framework for spiking neural network systems,'' \emph{arXiv preprint arXiv:2402.11322}, 2024.

\bibitem{Ref_Dong_NASbench201_ICLR20}
X.~Dong and Y.~Yang, ``Nas-bench-201: Extending the scope of reproducible neural architecture search,'' in \emph{ICLR}, 2020.

\bibitem{Ref_Howard_MobileNet_arXiv17}
A.~Howard \emph{et~al.}, ``Mobilenets: Efficient convolutional neural networks for mobile vision applications,'' \emph{arXiv preprint arXiv:1704.04861}, 2017.

\bibitem{Ref_Lee_DSCNN_TCDS19}
C.~Lee \emph{et~al.}, ``Deep spiking convolutional neural network trained with unsupervised spike-timing-dependent plasticity,'' \emph{IEEE Trans. on Cognitive and Developmental Systems (TCDS)}, vol.~11, no.~3, pp. 384--394, 2019.

\end{thebibliography}
\end{spacing}

\end{document}